\documentclass[10pt,twocolumn,letterpaper]{article}

\usepackage[pagenumbers]{iccv} 

\usepackage{multirow}

\definecolor{iccvblue}{rgb}{0.21,0.49,0.74}
\usepackage[pagebackref,breaklinks,colorlinks,allcolors=iccvblue]{hyperref}

\def\paperID{10595} 
\def\confName{ICCV}
\def\confYear{2025}

\definecolor{codegray}{gray}{0.9}
\newcommand{\code}[2]{%
  \begingroup\setlength{\fboxsep}{1pt}%
  \colorbox{#1}{#2}%
  \endgroup
}
\newcommand{\isscore}{\code{orange!3}{Intersectional Sensitivity}}
\newcommand{\modelname}{\code{cyan!5}{BiasConnect}}

\title{\modelname: Investigating Bias Interactions in Text-to-Image Models }

\author{Pushkar Shukla$^{1}$
~~~~~
Aditya Chinchure$^{2}$
~~~~~
Emily Diana$^{3}$
~~~~~
Alexander Tolbert$^{4}$ \\
~~~~~
Kartik Hosanagar$^{5}$
~~~~~
Vineeth N. Balasubramanian$^{6}$
~~~~~
Leonid Sigal$^{2}$
~~~~~
Matthew A. Turk$^1$ \\
$^1$Toyota Technological Institute at Chicago ~~~~~ $^2$University of British Columbia \\ $^3$Carnegie Mellon University, Tepper School of Business ~~~~~ $^4$Emory University \\
$^5$University of Pennsylvania, The Wharton School ~~~~~ $^6$Indian Institute of Technology Hyderabad\\
{\tt\small \{pushkarshukla, mturk\}@ttic.edu}  ~~~~~
{\tt\small \{aditya10, lsigal\}@cs.ubc.ca}
}

\begin{document}
\maketitle
\begin{abstract}
The biases exhibited by Text-to-Image (TTI) models are often treated as if they are independent, but in reality, they may be deeply interrelated. Addressing bias along one dimension, such as ethnicity or age, can inadvertently influence another dimension, like gender, either mitigating or exacerbating existing disparities. Understanding these interdependencies is crucial for designing fairer generative models, yet measuring such effects quantitatively remains a challenge.
In this paper, we aim to address these questions by introducing \modelname, a novel tool designed to analyze and quantify bias interactions in TTI models. Our approach leverages a counterfactual-based framework to generate pairwise causal graphs that reveals the underlying structure of bias interactions for the given text prompt. Additionally, our method provides empirical estimates that indicate how other bias dimensions shift toward or away from an ideal distribution when a given bias is modified. Our estimates have a strong correlation (+0.69) with the interdependency observations post bias mitigation. We demonstrate the utility of \modelname\ for selecting optimal bias mitigation axes, comparing different TTI models on the dependencies they learn, and understanding the amplification of intersectional societal biases in TTI models.

\end{abstract}    
\begin{figure}[t]
  \centering
   \includegraphics[width=1.0\linewidth]{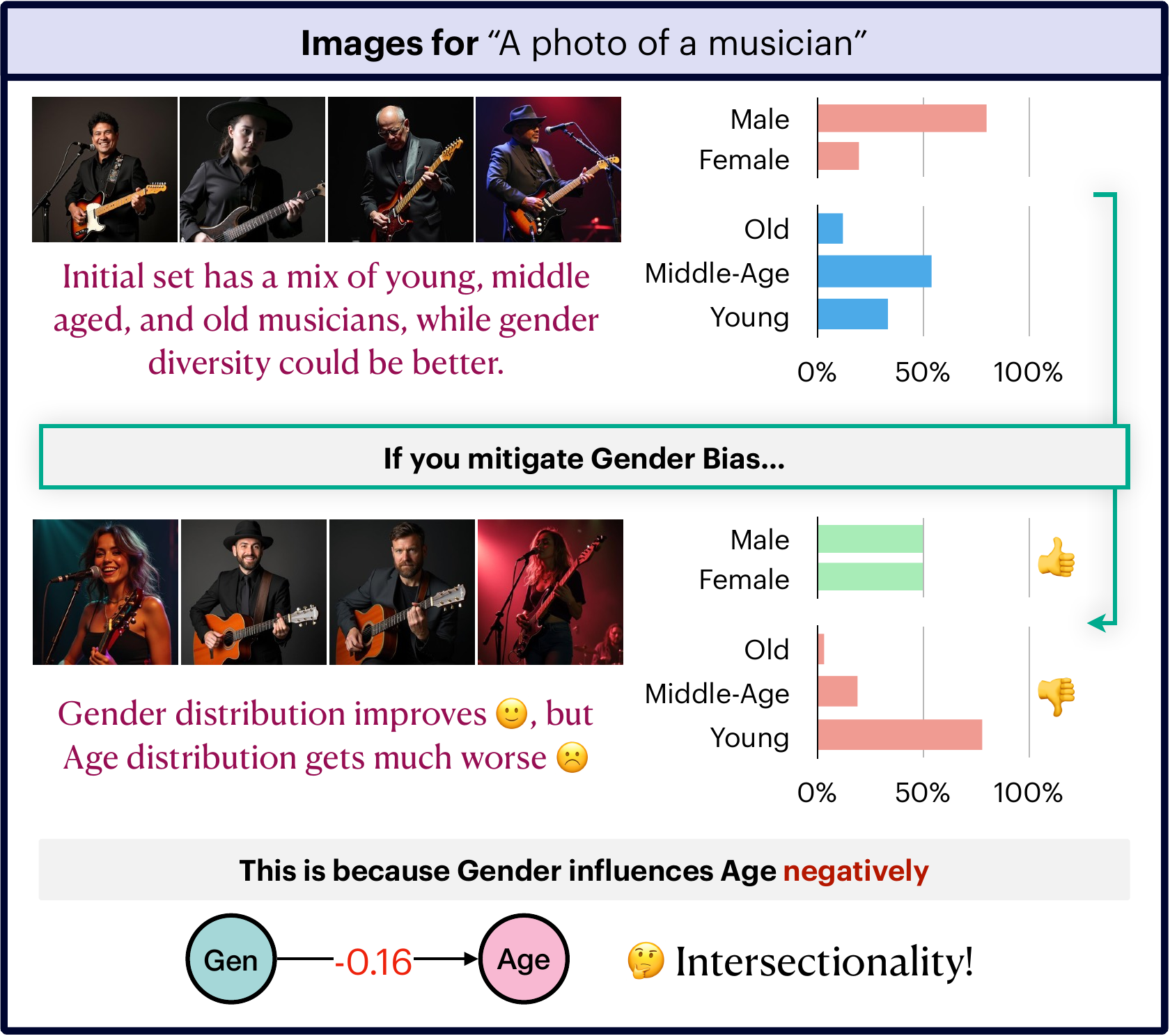}
   \caption{ An example output of \modelname, revealing the negative impact of bias mitigation along one dimension on another dimension. Here, increasing the gender diversity (GEN) skews age distribution (AGE) for images of musicians generated by Stable Diffusion 1.4 \cite{rombach2022high}. }
   \label{fig:header}
\end{figure}
\section{Introduction}
\label{sec:intro}

Text-to-Image (TTI) models such as DALL-E \citep{ramesh2021zero}, Imagen \citep{saharia2022photorealistic}, and Stable Diffusion \cite{rombach2022high} have become widely used for generating visual content from textual prompts. Despite their impressive capabilities, these models often inherit and amplify biases present in their training data \cite{wang_towards_2022,chinchure2024tibet,cho2023dall}. These biases manifest across multiple social  and non-social dimensions -- including 
gender, race, clothing, and age -- leading to skewed or inaccurate representations. As a result, TTI models may reinforce harmful stereotypes and societal norms \citep{bender2021dangers, birhane2021multimodal}. While significant efforts have been made to evaluate and mitigate societal biases in TTI models \citep{wang2023t2iat, cho2023dall,ghosh2023person,esposito2023mitigating,bianchi2023easily,chinchure2024tibet}, these approaches often assume that biases along different dimensions (e.g., gender and race) are independent of each other. Consequently, they do not account for relationships between these dimensions. For instance, as illustrated in Figure \ref{fig:header}, mitigating gender (male, female) may effectively diversify the gender distribution in a set of generated images, but this mitigation step may negatively impact the diversity of another bias dimension, like age. This relationship between two bias dimensions highlights the intersectional nature of these biases.

The concept of \textit{intersectionality}, first introduced by 
Crenshaw \citep{crenshaw1989demarginalizing}, motivates the need to understand how overlapping social identities such as race, gender, and class contribute to systemic inequalities. In TTI models, these intersections can have a significant impact. As a motivating study, we independently mitigated eight bias dimensions over 26 occupational prompts on Stable Diffusion 1.4, using a popular bias mitigation strategy, ITI-GEN \cite{zhang2023iti} (see Supp. \ref{sup:intro_study}). We found that while the targeted biases were reduced in most cases, biases along other axes were negatively affected in over $29.4\%$ of the cases. This suggests that for an effective bias mitigation strategy, it is crucial to understand which biases are intersectional. Additionally, it is important to consider whether mitigating one bias affects other biases equally or unequally, and if these effects are positive or negative.
Answering these questions can improve our understanding of TTI models, enhance their interpretability, and develop better bias mitigation strategies.

To understand how biases in TTI models influence each other, we propose \modelname, a first-of-its-kind analysis tool that evaluates societal biases in TTI models while accounting for intersectional relationships, rather than treating them in isolation. Our approach enables us to \textit{identify intersectional relationships} and \textit{study the positive and negative impacts} that biases have on each other, preventing unintended consequences that may arise from mitigating biases along a single axis. Our tool analyzes intersectionality at the prompt level, but also enables comparative studies across models by aggregating over a set of prompts, thus providing a means to uncover how variations in architecture, datasets, and training objectives contribute to bias entanglement. 

Given an input prompt to a TTI model, \modelname\ uses counterfactuals to quantify the impact of bias diversification (intervention) along one bias axis on any other bias axis. We refer to this as a pairwise causal relationship between the axis of intervention and the axis on which the effect is observed, and we visualize these relationships in the form of a \textit{pairwise causal graph}.
Additionally, we provide empirical estimates (called \isscore) that measure how bias mitigation along one dimension influences biases in other dimensions. These empirical estimates serve as weights on the pairwise causal graph.
To validate our approach, we show how our estimates correlate with true bias mitigation, and analyze robustness in which different components are systematically modified. Our overall contributions are as follows: 
\begin{itemize}
    \item We propose \modelname, a novel intersectional bias analysis tool for TTI models. Through a causal approach, our tool captures interactions between bias axes in a pairwise causal graph and provides empirical estimates of how bias mitigation along one axis affects other axes, through the \isscore\ score.    
    \item We show that our empirical estimates strongly correlate (+0.696) with the intersectionality observed post-mitigation, and through extensive qualitative results, validate the analyses provided by our tool.
    \item Finally, we demonstrate the usefulness of the tool in conducting audits on multiple open-source TTI models, identifying optimal mitigation strategies to account for intersectional biases, and showing how bias interactions in real-wold or a training dataset may change in TTI model generated images.
\end{itemize}

\section{Related Work }
\label{sec:related_work}
\subsection{Intersectionality and Bias in AI}

Intersectionality, introduced by Crenshaw \cite{crenshaw1989demarginalizing}, describes how multiple forms of oppression—such as racism, sexism, and classism—intersect to shape unique experiences of discrimination. Two key models define this concept: the additive model, where oppression accumulates across marginalized identities, and the interactive model, where these identities interact synergistically, creating effects beyond simple accumulation \cite{curry2018killing}. In the context of AI, most existing work \cite{diana2023correcting,kavouras2023fairness,kearns2018preventing, pmlr-v80-hebert-johnson18a} aligns more closely with the additive model, focusing on quantifying and mitigating biases in intersectional subgroups. This perspective has influenced fairness metrics \cite{diana2021minimax,foulds2020intersectional,ghosh2021characterizing} designed to assess subgroup-level performance, extending across various domains, including natural language processing (NLP) \cite{lalor2022benchmarking,lassen2023detecting,guo2021detecting,tan2019assessing} and recent large language models \cite{kirk_bias_2021,ma2023intersectional,devinney2024we,bai2025explicitly}, multimodal research \cite{howard2024socialcounterfactuals,hoepfinger2023racial}, and computer vision \cite{wang2020towards, steed2021image}. These approaches typically measure disparities across predefined demographic intersections and propose mitigation strategies accordingly. Our work aligns with the interactive model of intersectionality, using counterfactual-driven causal analysis in TTI models. Beyond subgroup analysis, we intervene on a single bias axis to assess its ripple effects on others, revealing independences and interactions.




\subsection{Bias in Text-to-Image Models}

Extensive research has been conducted on evaluating and mitigating social biases in both image-only models \cite{buolamwini2018gender, seyyed2021underdiagnosis, hendricks2018women, meister2023gender, wang2022revise, liu2019fair, joshi2022fair, wang2020towards, wang2023overwriting} and text-only models \cite{bolukbasi2016man, hutchinson2020social, shah2020predictive, garrido2021survey, ahn2021mitigating}. More recently, efforts have expanded to multimodal models and datasets, addressing biases in various language-vision tasks. These investigations have explored biases in embeddings \cite{hamidieh2023identifying}, text-to-image (TTI) generation \cite{cho2023dall, bianchi2023easily, seshadri2023bias, ghosh2023person, zhang2023iti, wang2023t2iat, esposito2023mitigating}, image retrieval \cite{wang2022assessing}, image captioning \cite{hendricks2018women, zhao2021scaling}, and visual question-answering models \cite{park2020fair, aggarwal2023fairness, hirota2022gender}.

Despite these advances, research on intersectional biases in TTI models remains limited. Existing evaluation frameworks such as T2IAT \cite{wang2023t2iat}, DALL-Eval \cite{cho2023dall}, and other studies \cite{ghosh2023person, esposito2023mitigating, bianchi2023easily, friedrich2023fair} primarily assess biases along predefined axes, such as gender \cite{wang2023t2iat, cho2023dall, ghosh2023person, esposito2023mitigating, bianchi2023easily}, skin tone \cite{wang2023t2iat, cho2023dall, ghosh2023person, esposito2023mitigating, bianchi2023easily}, culture \cite{esposito2023mitigating, wang2023t2iat}, and geographical location \cite{esposito2023mitigating}. 
While these works offer key insights into single-axis bias detection and mitigation, they lack a systematic examination of how biases on one axis influence another—a core aspect of intersectionality. The closest research, TIBET \cite{chinchure2024tibet}, visualizes such interactions, but our approach goes further by systematically quantifying bias interactions and empirically estimating their impact rather than merely identifying correlations.

\section{Approach}
\label{sec:Approach}

\begin{figure*}
  \centering
   \includegraphics[width=\linewidth]{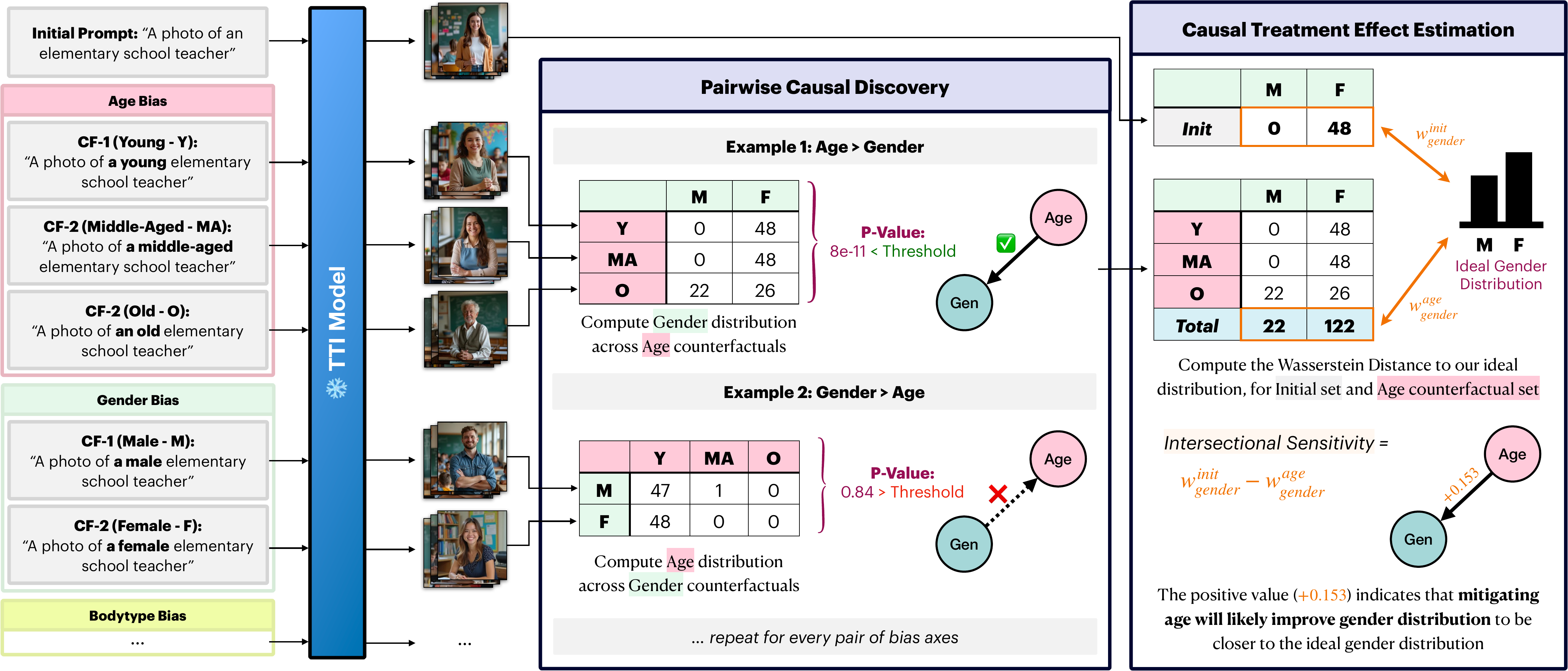}
   \caption{\textbf{An overview of \modelname}. We use a counterfactual-based approach to measure pairwise causality between bias axes. For dependent axes, we measure the causal effect, estimating how bias mitigation on one axis impacts another.}
   \label{fig:approach}
\end{figure*}

The objective of \modelname\ is to identify and quantify the intersectional effects of intervening on one bias axis (\(B_x\)) to mitigate that bias, on any other bias axis (\(B_y\)). \modelname, works by systematically altering input prompts and analyzing the resulting distributions of generated images. To achieve this, we leverage counterfactual prompts by modifying specific attributes (e.g., male and female) along a bias axis (e.g., gender) and examine how these interventions impact other bias dimensions (e.g., age and ethnicity).  If modifying one bias axis through counterfactual intervention causes significant shifts in the distribution of attributes along another bias axis, it indicates an intersectional dependency between these axes. 

We first construct prompt counterfactuals and generate images using a TTI model (Sec. \ref{sec:Approach:Imggen}). Subsequently, to identify bias-related attributes in the generated images, we use a VQA model (Sec. \ref{sec:Approach:VQA}). Next, in order to identify whether the intersectional effects of intervening on one bias on another axis is significant, we propose a causal discovery approach, where we employ conditional independence testing (Sec. \ref{sec:Approach:CausalDiscovery}) in a pairwise manner between the two bias axes. Finally, to quantify the intersectional effects, and to identify whether these effects are positive or negative, we compute the causal treatment effect, defined as \isscore\ (Sec. \ref{sec:Approach:CausalEffect}). 

\subsection{Counterfactual Prompts \& Image Generation}
\label{sec:Approach:Imggen}

Given an input prompt \( P \) and bias axes \( B = [B_1, B_2, \dots, B_n] \), we generate counterfactual prompts 
\(\{CF_i^1, \dots, CF_i^j\} \) using templates from Supp. \ref{sup:dataset}. The original prompt \( P \) and its counterfactuals are then used to generate images with the TTI model to measure intersectional effects.

\subsection {VQA-based Attribute Extraction}
\label{sec:Approach:VQA}
 
To facilitate the process of extracting bias related attributes from the generated images, we use VQA. This is inspired by previous approaches on bias evaluation, like TIBET \cite{chinchure2024tibet} and OpenBias \cite{d2024openbias}, where a VQA-based method was used to extract concepts from the generated images. Similar to previous work \cite{chinchure2024tibet}, we use MiniGPT-v2 \cite{chen2023minigptv2} in a question-answer format to extract attributes from generated images.

For the societal biases we analyze, we have a list of predefined questions (Supp. \ref{sup:VQA}) corresponding to each bias axis in $B$, and each question has a choice of attributes to choose from. For example, for the gender bias axis, we ask the question ``\texttt{\small [vqa] What is the gender (male, female) of the person?}''. Note that every question is multiple choice (in this example, \texttt{\small male} and \texttt{\small female} are the two attributes for gender). The questions asked for all images of prompt \( P \) and its counterfactuals \( CF_i^j \) remain the same. With the completion of this process, we have attributes for all images, where each image has one attribute for each bias axis in $B$.

\subsection{Pairwise Causal Discovery}
\label{sec:Approach:CausalDiscovery}
Given an initial set of bias axes \( B \), we define an intersectional relationship between a pair of biases \((B_x, B_y)\) as \( B_x \to B_y \), indicating that a counterfactual intervention on \( B_x \) to mitigate its bias also affects \( B_y \). As a first step, we intervene across all \( n \times n \) bias relationships. Using attributes extracted by the VQA, we can count the attributes for a bias axis $B_y$ over any set of images. We construct a contingency table where rows represent the intervened bias axis \( B_x \) (e.g., gender with male and female counterfactuals, in Example 2 of Fig. \ref{fig:approach}), and columns capture the distribution on the target axis \( B_y \) (e.g., age with old, middle-aged, and young categories). The values in the contingency tables are the counts of attributes of $B_y$ over the counterfactual image sets of $B_x$.


Next, we refine these relationships by extracting only statistically significant ones. This ensures that only strong dependencies between different bias pairs are retained. We apply conditional independence testing using the Chi-square (\(\chi^2\)) test, pruning bias pairs with respect to \( B_x \) if their p-value exceeds a predefined threshold (p-value$>0.0001$). Bias pairs with a p-value below this threshold are considered strongly dependent, indicating that intervening on \( B_x \) results in a significant change in the other bias axis. This process is applied iteratively for all bias axes. This step is referred to as \textit{\textbf{Pairwise Causal Discovery}}, and it returns a set of bias pair relationships where mitigating along one bias axis has led to a strong change in another bias dimension

\begin{figure*}[ht]
  \centering
   \includegraphics[width=\linewidth]{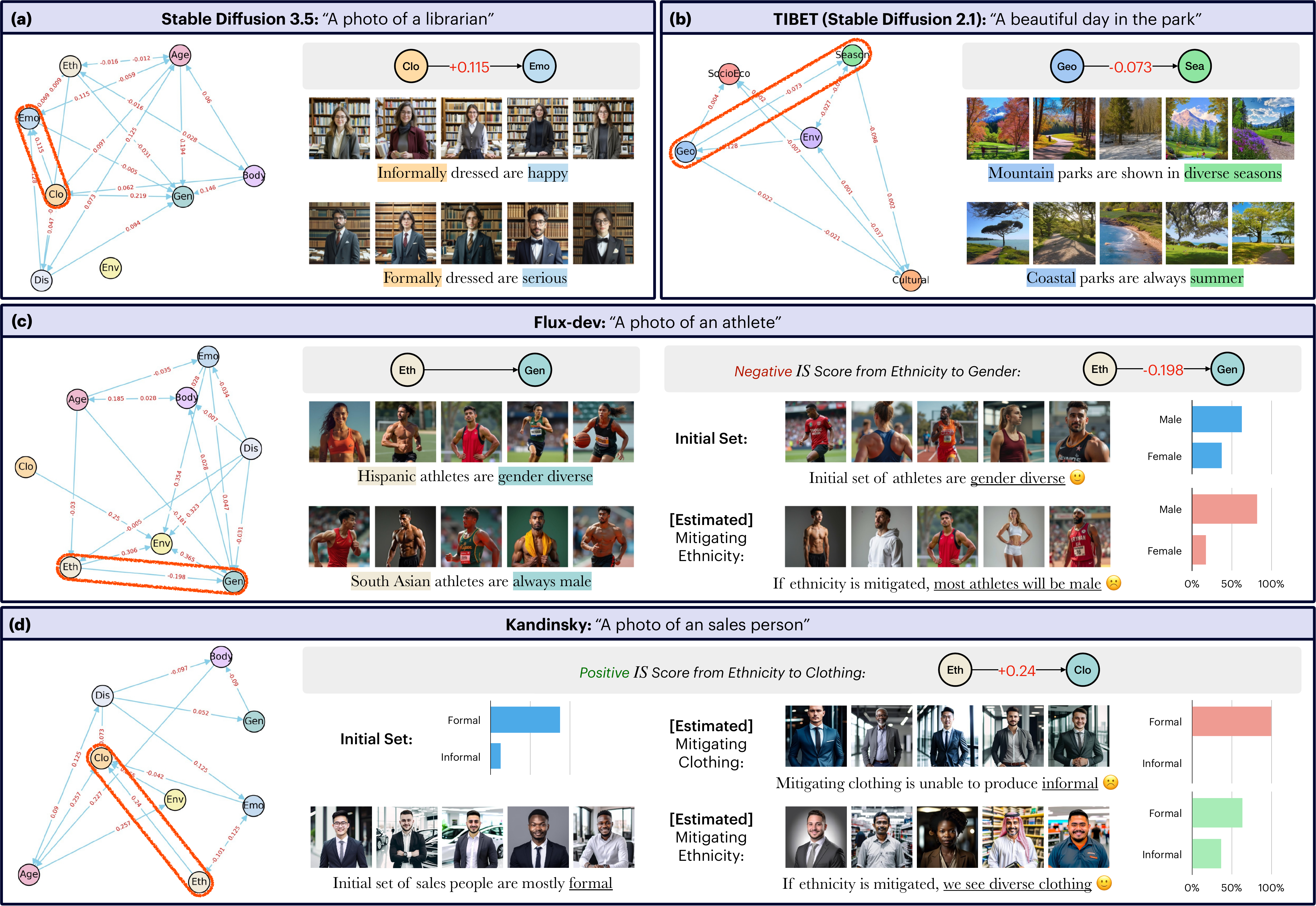}
   \caption{The figure illustrates bias interpretations from Bias Connects, combining all pairwise graphs into one. (a) Shows how mitigating clothing bias also mitigates emotion bias. (b) Explores interactions between non-traditional bias axes in the TIBET dataset. (c) Reveals that generating ethnically diverse athletes reduces gender diversity. (d) Demonstrates that diversifying salesperson clothing is best achieved by increasing ethnic diversity rather than directly specifying clothing variation. }
   \label{fig:examples}
\end{figure*}

\subsection{Causal Treatment Effect Estimation}
\label{sec:Approach:CausalEffect}
While Pairwise Causal Discovery can identify interventions along bias pairs that cause significant changes, it alone is not sufficient to determine whether the impact of interventions on \( B_x \) affects \( B_y \) in a positive or negative direction with respect to an ideal distribution. This limitation arises because there is no direct comparison to the initial distribution of \(B_y\) in the original set of images generated from prompt \(P\), as we had only considered images from the counterfactuals $\{CF_x^1,...,CF_x^j\}$ for bias $B_x$, for pairwise causal discovery. To address this, we propose a metric that quantifies the impact of bias mitigation on dependent biases with respect to an ideal distribution. 



\noindent{\bf Defining an Ideal Distribution.} We first define a desired (ideal) distribution \(D^*\), which represents the unbiased state we want bias axes to achieve. This can be a real-world distribution of a particular bias axis, a uniform distribution (which we use in our experiments), or anything that suits the demographic of a given sub-population. 

\noindent{\bf Measuring Initial Bias Deviation.} Given the images of initial prompt \(P\), we compute the empirical distribution of attributes associated with bias axis \(B_y\), denoted as \( D_{B_y}^{\text{init}} \). We then compute the Wasserstein distance between this empirical distribution and the ideal distribution:
\begin{equation}
w_{B_y}^{\text{init}} = W_1(D_{B_y}^{\text{init}}, D^*)
\end{equation}
\noindent where \( W_1(\cdot, \cdot) \) represents the Wasserstein-1 distance. The Wasserstein-1 distance (also known as the Earth Mover's Distance) between two probability distributions \( D_1 \) and \( D_2 \) is defined as:
\begin{equation}
W_1(D_1, D_2) = \inf_{\gamma \in \Pi(D_1, D_2)} \mathbb{E}_{(x,y) \sim \gamma} [|x - y|]
\end{equation}
\noindent where \( \Pi(D_1, D_2) \) is the set of all joint distributions \( \gamma(x, y) \) whose marginals are \( D_1 \) and \( D_2 \), and \( |x - y| \) represents the transportation cost between points in the two distributions.

\noindent{\textbf{Intervening on $B_x$.}}
Next, we intervene on \(B_x\) to simulate the mitigation of bias $B_x$. This intervention ensures that all counterfactuals of \(B_x\) are equally represented in the generated images. For example, if \(B_x\) is gender bias, we enforce equal proportions of male and female individuals in the dataset. This intervention is in line with most bias mitigation methods proposed for TTI models, like ITI-GEN \cite{zhang2023iti}. Using our counterfactuals along \(B_x\), we sum the distributions on \(B_y\) across all counterfactuals of \(B_x\). This sum across the counterfactuals of \( B_x \) yields a new empirical distribution of \( B_y \), denoted \( D_{B_y}^{B_x} \), simulating the effect of mitigating \( B_x \) (See Fig \ref{fig:approach}). We compute its Wasserstein distance from the ideal distribution.
\begin{equation}
w_{B_y}^{B_x} = W_1(D_{B_y}^{B_x}, D^*)
\end{equation}
\noindent{\textbf{Computing \isscore.}} To quantify the effect of mitigating \(B_x\) on \(B_y\), we define the metric, \isscore, as:
\begin{equation}
IS_{xy} = w_{B_y}^{\text{init}} - w_{B_y}^{B_x} 
\label{eqn:metric}
\end{equation}
\noindent A positive value (\( IS_{xy} > 0 \)) indicates that mitigating \( B_x \) improves \( B_y \), bringing it closer to the ideal distribution, while a negative value (\( IS_{xy} < 0 \)) suggests it worsens \( B_y \), moving it further from the ideal. If \( IS_{xy} = 0 \), mitigating \( B_x \) has no effect on \( B_y \). This approach enables us to assess whether addressing one bias (e.g., gender) improves or worsens another (e.g., ethnicity) in generative models, providing a systematic way to evaluate trade-offs and unintended consequences in bias mitigation strategies. We use \isscore\ ($IS_{xy}$) as a measure of intersectionality for \(B_x \to B_y\).


\subsection{Visualization}

Following the process above, we have a set of pairwise causal relationships for all significant intersectional bias pairs $B_x \to B_y$. Furthermore, each pair $B_x \to B_y$ has an \isscore\ score to quantify the intersectional effects. There are many ways to represent these pairwise relationships, including building an $n \times n$ matrix, or a graph with $n$ nodes and directed edges that represent the relationships between these nodes. 

A user of \modelname\ may want to understand all important  intersectional effects together. To that end, we adopt a graph representation for our output. This graph is referred to as a \textbf{Pairwise Causal Graph} in the rest of the paper. Figures \ref{fig:examples} and \ref{fig:main} show examples of such graphs. To interpret this graph, first pick a focal node where the intervention takes place. All outgoing edges from this node indicate intersectional relationships that are statistically significant. The weights of the edges show the \isscore\ and can be interpreted as the impact of intervention on the bias axis for the focal node.

\section{Causal Interpretations}


\noindent{\bf Pairwise Causal Discovery.}
Our approach is causal as it involves explicit interventions to measure the effect of one variable on another, aligning with Pearl’s Ladder of Causality \cite{pearl2009causality}. Rather than analyzing existing images, we actively modify bias attributes (e.g., gender, race, age) in input prompts. However, our pairwise causal discovery pipeline does not capture indirect causal effects between bias axes.

\noindent{\textbf{Causal Treatment Effect.}} The \isscore\ metric is a \textit{causal treatment effect} metric because it quantifies how mitigating one bias ($B_x$) causally influences another bias ($B_y$) through an intervention-based approach. By actively modifying $B_x$ (e.g., ensuring equal representation across its attributes) and measuring changes in the Wasserstein distance of $B_y$ from an ideal distribution, we estimate the causal impact of debiasing. This aligns with \textit{counterfactual causal inference} \cite{morgan2014counterfactuals}, where we compare the observed outcome ($B_y$ distribution) with its initial state had no intervention occurred. The method follows \textit{Rubin's causal model}  \cite{rubin1974estimating,rubin1975bayesian}, treating bias mitigation as a \textit{treatment-control experiment}, and can be represented in a \textit{Directed Graph} as $B_x \to B_y$, making it distinct from mere correlation analysis. The metric $IS_{xy}$ in Equation \ref{eqn:metric} captures the magnitude of causal influence, providing insights into whether mitigating one bias improves or worsens another.

\section{Experiments}
\label{sec:Experiments}

In this section, we begin by explaining the two datasets—the occupation prompts and the TIBET dataset—that we use to test \modelname\ (Sec. \ref{sec:ModelsDatasets}). Following that, show the usefulness of \modelname\ by analyzing prompts to study prompt-level bias intersectionality (Sec. \ref{sec:mainintersectionality}), and validate our \isscore\ with the help of a downstream bias mitigation tool, ITI-GEN (Sec. \ref{sec:mitigation}). Finally, we analyze the robustness of \modelname\ on the number of images generated per prompt, and errors in VQA (Sec. \ref{sec:robustness}).

\subsection{Models and Datasets}
\label{sec:ModelsDatasets}


\noindent\textbf{Occupation Prompts.}  To facilitate a structured evaluation, we develop a dataset with 26 occupational prompts, along eight distinct bias dimensions: gender, age, ethnicity, environment, disability, emotion, body type, and clothing. We generate 48 images for all initial counterfactual prompts using five Text-to-Image models: Stable Diffusion 1.4, Stable Diffusion 3.5, Flux \cite{flux2024}, Playground v2.5 \cite{li2024playground} and Kandinsky 2.2 \cite{kandinsky22,razzhigaev2023kandinsky}. Further details about the prompts, bias axes, and counterfactuals are provided in the Supp. \ref{sup:dataset}.

\noindent\textbf{TIBET dataset.} 
The TIBET dataset includes 100 creative prompts with LLM-generated bias axes and counterfactuals \cite{chinchure2024tibet}. Its diversity of prompts and bias axes, unrestricted to a fixed set, enhances its utility. Additionally, it provides 48 Stable Diffusion 2.1-generated images per initial and counterfactual prompt (See Supp. \ref{sup:TIBET} for more details).

\subsection{Studying prompt-level intersectionality}
\label{sec:mainintersectionality}

\modelname\ enables prompt-level analysis of intersectional biases (Fig. \ref{fig:examples}), helping users identify key bias axes and develop effective mitigation strategies. For instance, in Fig. \ref{fig:examples}(a), Stable Diffusion 3.5 exhibits a causal link between clothing and emotion bias—informally dressed librarians appear happy, while formally dressed ones seem serious. A strongly positive \isscore\ ($IS$= 0.115) indicates that diversifying clothing alone is sufficient to diversify emotion, without explicitly mitigating emotion bias. Conversely, Fig. \ref{fig:examples}(c) illustrates how ethnicity can negatively impact gender diversity. South Asian athletes, for example, are predominantly depicted as male. The negative \isscore\ ($IS = -0.198$) suggests that mitigating ethnicity alone would further skew gender representation toward males. These interpretations of our tool have various applications, including identifying optimal bias mitigation strategies and comparing multiple TTI models, as discussed in Section \ref{sec:Applications}. 


\subsection{Validating \isscore}
\label{sec:mitigation}
Our approach estimates how counterfactual-based mitigation affects bias scores using the \isscore. To validate this, we debias ITI-GEN, mitigate biases along each dimension, and measure the correlation between pre- and post-mitigation \isscore\ values. As shown in Table \ref{tab:mitigationcorr}, we achieve an average correlation of +0.696 across occupations, with higher values for specific prompts like Nurse (+0.997). The strong correlation observed between pre- and post-mitigation bias scores suggests that our approach effectively captures the potential impacts of interventions, offering a transparent and data-driven way to evaluate model fairness. More details regarding our experimental setup have been provided in Supp. \ref{sup:mitigation}.

\begin{table}[t]
  \small
  \centering
  \begin{tabular}{lcccc}
    \toprule
    \textbf{Prompt} & \textbf{Edges} & \textbf{Corr.} & \textbf{MaxInf} & \textbf{MaxImp}\\
    \midrule
    Pharmacist  &  12  &  +0.399 & Gender & Age\\
    Scientist  &  9  &  +0.600 & Clothing & Ethnicity\\
    Doctor  &  9  &  +0.638 & Age & Disability \\
    Librarian  &  14  &  +0.805 & Emotion & Age \\
    Nurse  &  5  &  \underline{+0.997}  & Age & Disability\\
    Chef  &  8  &  +0.757 & Bodytype & Ethnicity\\
    Politician  &  10  &  +0.782 & Emotion & Disability \\
    \midrule
    \textbf{Overall} & - & \textbf{+0.696} & Gender & Age\\
    \bottomrule
  \end{tabular}
  \caption{\textbf{Correlation Between Estimates and Post-Mitigation Evaluation on ITI-GEN.} The high correlation validates our mitigation estimates. For each prompt, we report one of the most influenced node (MaxInf) and the node with the greatest impact on others (MaxImp).
 }
  \label{tab:mitigationcorr}
\end{table}

\subsection{Robustness of \modelname}
\label{sec:robustness}

We analyze the robustness of our method by evaluating the impact of image generation and VQA components on pairwise causal graphs and \isscore\ values through experiments on image set size and VQA error rates across occupation prompts. This robustness analysis is useful because it ensures the reliability and stability of our method across varying conditions.

\begin{figure}
  \centering
   \includegraphics[width=\linewidth]{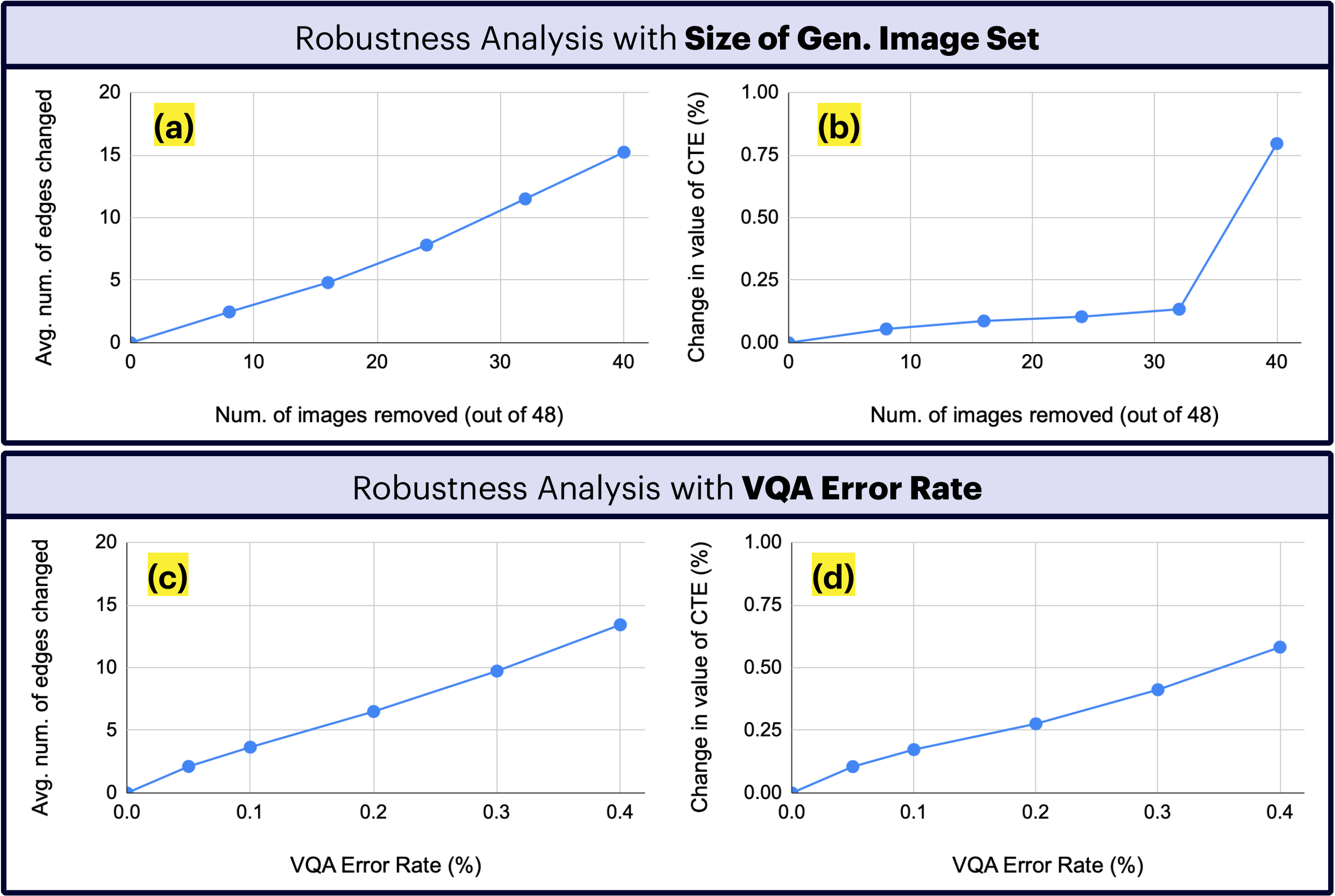}
   \caption{\textbf{Sensitivity analysis on \modelname}. . We evaluate the robustness of our approach by analyzing the impact of VQA errors and the effect of the number of images on the pairwise causal graph and \isscore.}
   \label{fig:sensitivity}
\end{figure}

\noindent\textbf{Number of Images.} Our method generates 48 images per prompt to study bias distributions reliably. To assess the impact of reducing image count, we analyze changes in the total number of edges in the pairwise causal graph and the percentage change in \isscore\ (Fig. \ref{fig:sensitivity}(a-b)). Removing 8 images (16.6$\%$) results in only 2.4 edge changes and a minor 5.5$\%$ shift in \isscore. Even with 16 images removed (33.3$\%$), only 4.8 edges change, and \isscore\ shifts by 8$\%$. This low impact suggests that TTI models consistently generate similar bias distributions (e.g., always depicting nurses as females), preserving overall trends despite fewer images. However, excessive pruning significantly affects the analysis—removing 40 images (83$\%$) leads to a sharp 79$\%$ change in \isscore. This demonstrates that our approach is robust to moderate reductions in image count but breaks down when the sample size is too small. While a sufficiently large image set enhances reliability, exceeding 48 images offers only marginal analytical benefits.

\noindent\textbf{VQA Error Rate.} In Fig. \ref{fig:sensitivity}(c-d), we show the impact of VQA errors on the graph and \isscore\ values. We randomly change the VQA answers to a different answer (simulating an incorrect answer) at different thresholds, from 5\% to 40\% of the time. We observe that with low error rates of 5\% and 10\%, the impact on number of edges changed is low, with averages of 2.1 and 3.65 edges respectively. However, the small changes in VQA answers does impact \isscore\ values, at 10\% and 17.3\% respectively, as the impact is compounded by the fact that we use both the initial and the counterfactual distributions to obtain this value, and that a 5\% error causes 13,478 answers out of a total of 269,568 answers to be changed, which is substantial. 
Nonetheless, we note that this impact remains linear. Our study shows the graph is more robust if the error rate is below 20$\%$. As VQA models improve, achieving error rates for robust graphs becomes practical. 

\section{Applications}
\label{sec:Applications}


\subsection{Applying \modelname\ to analyze TTI models}
\label{sec:globalanalysis}

\begin{figure*}
  \centering
   \includegraphics[width=\linewidth]{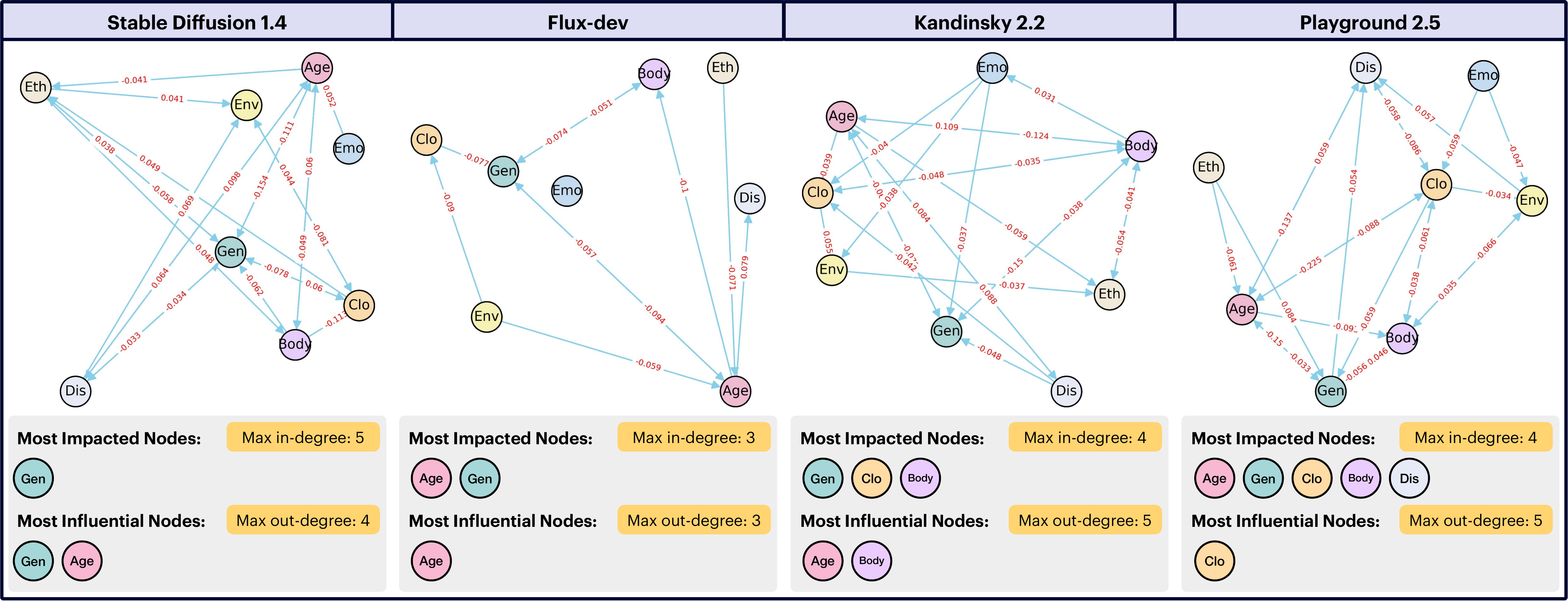}
   \caption{We compare aggregated causal graphs for four models: Stable Diffusion 1.4, Flux-dev, Kandinsky 2.2, and Playground 2.5. These graphs combine pairwise causal relationships across all bias axes, accumulated from occupation prompts in our dataset.}
   \label{fig:main}
\end{figure*}

\begin{figure}
  \centering
   \includegraphics[width=0.8\linewidth]{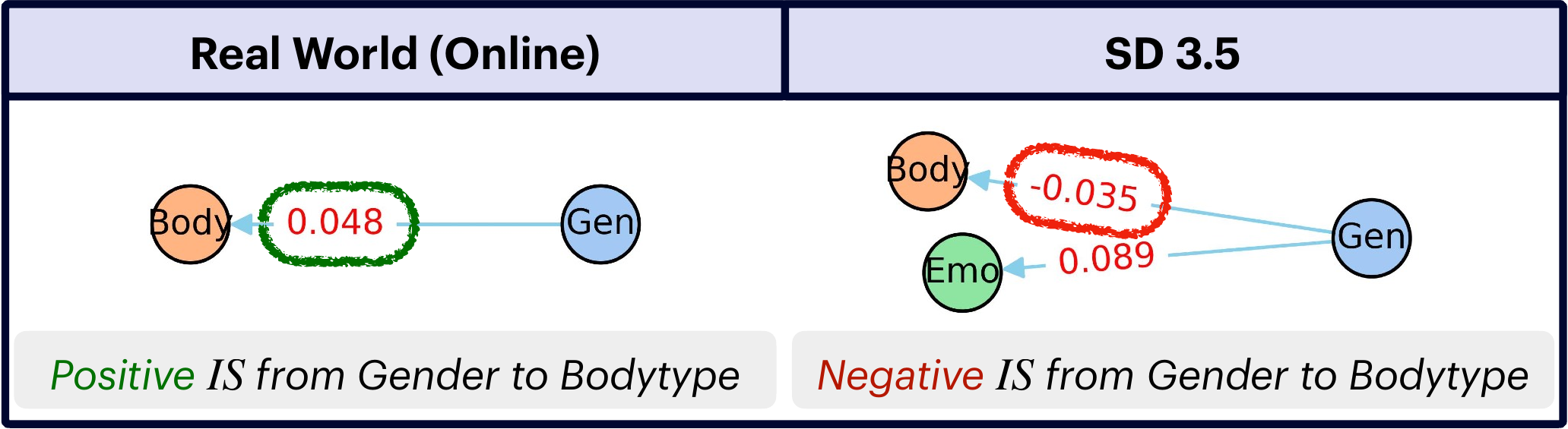}
    \caption{Comparison of real-world and Stable Diffusion 3.5 pairwise causal relationships for gender in computer programmer images. In the real world, gender diversification increases body type diversity, whereas in Stable Diffusion 1.4, it has a negative impact. }
   \label{fig:realworld}
\end{figure}

To compare bias interactions across models, we aggregate results from all prompts to create a unified representation, enabling a high-level analysis of bias trends (Fig. \ref{fig:main}). Details on the aggregation process are in the Supp. \ref{sup:aggregation}.

\noindent\textbf{Identifying high-impact biases.} Some biases act as primary sources, influencing multiple others, while some function as effects, shaped by upstream factors. A node’s impact is measured by its outgoing edges (\textbf{MaxImp}, Table \ref{tab:mitigationcorr}), while its susceptibility to influence is quantified by incoming edges (\textbf{MaxInf}). This helps in model selection based on specific bias priorities.

As an example, let's analyze how this information can help in selecting appropriate models using the global graphs in Figure \ref{fig:main}. If a user prioritizes robustness to age-related bias when selecting a model, Kandinsky 2.2 would be the best choice, as its Age node is the least influenced by other biases in the global analysis. This means that modifying other attributes (e.g., gender or clothing) has minimal unintended effects on age representation, ensuring more stable and independent age depictions across generated images.

Similarly, if the goal is to generate occupation-related images while minimizing unintended bias propagation across other attributes, Playground 2.5 is the optimal choice. In this model, variations in body type have the least impact on other biases, meaning changes in body shape do not disproportionately affect other attributes like gender, ethnicity, or perceived professionalism. This makes Playground 2.5 preferable in scenarios where maintaining fairness across multiple dimensions while altering body type is critical. By analyzing bias influence and susceptibility, users can make informed choices based on fairness priorities, whether aiming for stability in a bias axis or minimizing unintended shifts in related attributes.

\subsection{Studying Real-World Biases}
\label{sec:real_world_biases}
\modelname\ can also be used to compare the distribution of images generated by TTI models with real-world data. To demonstrate this, we sampled 48 images of computer programmers, and 48 images of male and female computer programmers each from the internet. We then compared the pairwise causal graph for gender in the real-world distribution to the one generated by Stable Diffusion 3.5. Our analysis (Fig. \ref{fig:realworld}) reveals that in real-world data, gender diversification primarily influences body type in a positive manner. However, in Stable Diffusion 3.5, gender impacts both emotion and body type, with a negative effect on body type, suggesting that increasing gender diversity reduces body type diversity. Studies like these are valuable in identifying discrepancies between generated images and real-world distributions or training datasets, and show how real-wold bias interactions may be amplified in TTI models. Supp \ref{sup:real_world} has further details on this process.

\subsection{Uncovering Optimal Bias Mitigation Strategies}
\label{sec:mitigationstrategy}
\modelname\ quantifies the impact of one bias on another, helping identify effective bias mitigation strategies for a given prompt. We illustrate this with three examples:

\noindent\textbf{Clothing and Emotion Bias (Fig. \ref{fig:examples}(a))} – Stable Diffusion 3.5 exhibits a causal link where informally dressed librarians appear happy, while formally dressed ones seem serious. A positive \isscore\ ($IS$ = 0.115) suggests that diversifying clothing alone is sufficient to diversify emotion without explicitly addressing emotion bias.

\noindent\textbf{Ethnicity and Gender Bias (Fig. \ref{fig:examples}(c))} – South Asian athletes are predominantly depicted as male. A negative \isscore\ ($IS$ = -0.198) indicates that mitigating ethnicity alone would further skew gender representation toward males. A better approach would involve fine-tuning on a dataset with more female South Asian athletes to improve disentanglement between ethnicity and gender.

\noindent\textbf{Ethnicity and Clothing Diversity (Fig. \ref{fig:examples}(d))} – For salespersons, the best way to diversify clothing styles is not by directly mitigating clothing bias but by increasing ethnic diversity in generated images. This reveals a hidden mitigation strategy where altering one axis (ethnicity) impacts another (clothing) more than direct intervention.


These examples highlight a key utility of \modelname: enabling users to adopt complementary bias mitigation strategies based on their specific needs. In some cases, mitigating one bias naturally diversifies another, reducing the need for direct intervention. In other cases, addressing one axis may worsen another, requiring a more targeted approach. Finally, certain biases may be best mitigated indirectly by adjusting a different, more influential axis.

\section{Conclusion}
\label{sec:Conclusion}
Our study proposes a tool to investigate intersectional biases in TTI models. While prior research has explored bias detection and mitigation in generative models, to the best of our knowledge, no previous work has focused on understanding how biases influence one another. We believe our work makes a significant contribution by enabling a more nuanced analysis of bias interactions.
Beyond academic research, \modelname\ has practical applications, including comparing biases dependencies learned across different models, establishing empirical guarantees for mitigation, and determining optimal mitigation approaches that account for intersectionality. We hope that this tool will facilitate more informed decision-making for AI practitioners, policymakers, and developers, ultimately leading to more equitable and transparent generative models.

While \modelname\ provides a valuable framework, it represents only an initial step toward a more comprehensive causal approach to understanding intersectionality. Our current setup does not allow us to reason about indirect causal effects, or develop an optimal bias mitigation strategy that utilizes our tool to mitigate multiple biases simultaneously. Addressing these challenges presents an important avenue for future research.


\noindent{\textbf{Ethical Considerations.}}
We acknowledge that the presence of biases in generative AI models can lead to real-world harms, reinforcing stereotypes and disproportionately affecting marginalized groups. Our tool is intended to provide researchers and practitioners with a means to better understand and mitigate these biases, rather than to justify or amplify them. Additionally, we recognize that bias analysis can be sensitive to the choice of datasets, evaluation methods, and experimental assumptions, and we encourage future work to refine and expand upon our approach.

{
    \small
    \bibliographystyle{ieeenat_fullname}
    \bibliography{main}
}
\clearpage
\setcounter{page}{1}
\maketitlesupplementary
\renewcommand{\thesection}{\Alph{section}}
\setcounter{section}{0}

\section{Supplementary Material}
\label{sec:sup}

\subsection{Occupation Dataset}
\label{sup:dataset}

Table \ref{tab:fulldataset} shows the occupation prompt templates along each of the bias axes. We consider 26 occupations: \texttt{\small computer programmer, elementary school teacher, librarian, announcer, pharmacist, chef, chemist, police, accountant, architect, lawyer, philosopher, scientist, doctor, nurse, engineer, musician, journalist, athlete, social worker, sales person, politician, farmer, mechanic, firefighter, gardener}.
\begin{table}
  \centering
  \resizebox{\columnwidth}{!}{
  \begin{tabular}{@{}ll@{}}
    \toprule
     \textbf{Axis} & \textbf{Counterfactuals} \\
     \midrule
\multirow{2}{*}{gender bias}  &  A photo of a male \textit{occupation} \\
 &  A photo of a female \textit{occupation} \\
\midrule
\multirow{3}{*}{age bias}  &  A photo of an old \textit{occupation} \\
  &  A photo of a middle-aged \textit{occupation} \\
 &  A photo of a young \textit{occupation} \\
\midrule
\multirow{6}{*}{ethnicity bias} &  A photo of a white \textit{occupation} \\
 &  A photo of a african american \textit{occupation} \\
  &  A photo of an asian \textit{occupation} \\
 &  A photo of a south asian \textit{occupation} \\
  &  A photo of a middle eastern \textit{occupation} \\
  &  A photo of a hispanic \textit{occupation} \\
\midrule
\multirow{3}{*}{bodytype bias} &  A photo of a obese \textit{occupation} \\
  &  A photo of a slim \textit{occupation} \\
  &  A photo of a muscular \textit{occupation} \\
\midrule
\multirow{2}{*}{environment bias}  &  A photo of a \textit{occupation} working indoors \\
 &  A photo of a \textit{occupation} working outdoors \\
\midrule
\multirow{2}{*}{clothing bias}  &  A photo of a \textit{occupation} in formal attire \\
 &  A photo of a \textit{occupation} in informal attire \\
\midrule
\multirow{4}{*}{emotion bias}  &  A photo of a \textit{occupation} who is happy \\
 &  A photo of a sad \textit{occupation} who is sad \\
  &  A photo of a \textit{occupation} who is serious \\
 &  A photo of a \textit{occupation} who is tired \\
\midrule
\multirow{4}{*}{disability bias}  &  A photo of a \textit{occupation} who is fit \\
 &  A photo of a blind \textit{occupation} \\
  &  A photo of a \textit{occupation} with a hearing aid \\
  &  A photo of a \textit{occupation} on a wheelchair \\
    \bottomrule
    \end{tabular}}
    \caption{\textbf{Occupation Dataset.}}
    \label{tab:fulldataset}
\end{table}

\subsection{Models}
\label{sup:models}

We test \modelname\ on several different TTI models. Unless mentioned otherwise, all the settings for image generation are the default settings provided for these models on Huggingface.

For prompts in the Occupation dataset, we use the following models:

\begin{itemize}
    \item Stable Diffusion 1.4: We pick this model because of its compatibility with most bias mitigation tools, like ITI-GEN. We use \textit{CompVis/stable-diffusion-v1-4}.
    \item Stable Diffusion 3.5: This is the latest model in the Stable Diffusion family. We specifically use the \textit{stabilityai/stable-diffusion-3.5-large-turbo} model.
    \item Flux-dev: The Flux series of models is another popular set of open-source TTI models. We use the \textit{black-forest-labs/FLUX.1-dev} variant of this family, and conduct inference with \texttt{guidance scale = 3.5} and \texttt{num inference steps = 30}.
    \item Playground 2.5: This model is trained to produce aesthetically pleasing images. We use \textit{playgroundai/playground-v2.5-1024px-aesthetic}, with \texttt{guidance scale = 3} and \texttt{num inference steps = 50}.
    \item Kandinsky 2.2: We use the \textit{kandinsky-community/kandinsky-2-2-decoder} model, and use the default \texttt{negative prompt = "low quality, bad quality"}.
\end{itemize}

For the TIBET dataset, we use the images already provided in the dataset as is. These images were generated using Stable Diffusion 2.1.

\subsection{VQA}
\label{sup:VQA}

For a given set of images and a set of axes $B$, our goal is to find distributions $D_{B_i}$ for all bias axes $B_i \in B$. In order to find this distribution, we must begin by first identifying the attributes related to $B_i$ in every image of the image set. We use VQA for this process.

For every image in the set, we first start by asking the VQA the question \texttt{\small Is there a person in the image (yes or no)?} for the Occupation prompts dataset. This allows us to filter images where we will be unable to extract bias-related attributes due to low quality generation. For the images that have a person, we have the following set of questions to extract all bias-related attributes:

\begin{itemize}
\item  gender bias:  \texttt{\small What is the gender (male, female) of the person?}
\item  age bias:  \texttt{\small What is the age group (young, middle, old) of the person?}
\item  ethnicity bias:  \texttt{\small What is the ethnicity (white, black, asian, south asian, middle eastern, hispanic) of the person?}
\item  bodytype bias:  \texttt{\small What is the body type (fat, slim, muscular) of the person?}
\item  environment bias:  \texttt{\small What is the environment (indoor, outdoor) of the person?}
\item  clothing bias:  \texttt{\small What is the attire (formal, informal) of the person?}
\item  emotion bias:  \texttt{\small What is the emotion (happy, sad, serious, tired) of the person?}
\item  disability bias:  \texttt{\small Is this person blind (yes or no)?; Is this person wearing a hearing aid (yes or no)?; Is this person on a wheelchair (yes or no)?}
\end{itemize}

Note that all questions are multiple choice. Furthermore, for disability bias, we split the question into three parts, and run each part through the VQA model independently. If none of the parts are answered as `yes', then the person in the image is `fit' and does not have one of those disabilities.

In terms of error rate for robustness, we believe that our MCQ-based VQA approach would yield a lower than 18\% error rate observed in TIBET \cite{chinchure2024tibet}, which uses the same VQA model. Empirically speaking, we observe that our VQA performs near-perfectly on axes such as gender, environment and emotion, but may sometimes return incorrect guesses among other axes in more ambiguous scenarios. As VQA models improve, our method can utilize them in a plug-and-play manner.

\subsection{TIBET Data}
\label{sup:TIBET}

TIBET dataset contains 100 prompts, their biases and relevant counterfactuals, and 48 images for each initial and counterfactual prompt. Because of the dynamic nature of these biases (they vary from prompt to prompt), we use the VQA strategy in the TIBET method \cite{chinchure2024tibet} instead of our templated questions from above. Moreover, in the causal discovery process, because tibet concepts are more diverse than the fixed attributes we use with occupation prompts, our p-value threshold changes to $0.05$.

\subsection{Bias Mitigation Study}
\label{sup:intro_study}

We conduct a study using ITI-GEN to measure how often a bias mitigation might yield negative effects on other bias axes. We define a negative \isscore\ score ($IS_{xy} < 0$) to suggest that mitigating bias axis $B_x$ reduces the diversity of attributes of axis $B_y$. 

In this study, for all 26 occupations and across all bias axes listed in Table \ref{tab:fulldataset}, we mitigate every bias axis independently. We then compute \isscore\, where the initial distribution $D_{B_y}^{B_x}$ in equation 3 is replaced by $D_{B_y}^{mit(B_x)}$, which is based on the VQA extracted attributes for bias axis $B_y$ in the newly generated set of images post-mitigation of axis $B_x$ with ITI-GEN. 
This score is defined as:
\begin{equation}
w_{B_y}^{B_x} = W_1(D_{B_y}^{mit(B_x)}, D^*)
\end{equation}
\begin{equation}
IS_{xy}^{mit(x)} = w_{B_y}^{\text{init}} - w_{B_y}^{B_x} 
\end{equation}

We compute the percentage of $IS_{xy}^{mit(x)}$ for all possible pairs of biases, $B_x$ and $B_y$, where mitigation of $B_x$ led to $IS_{xy}^{mit(x)} < 0$. We find that a substantial number of times, $29.4\%$ of all mitigations, led to a negative effect.

\subsection{Additional prompt-level examples}
\label{sup:TIBET}

We show additional examples of prompt-level intersectional analysis in Fig \ref{fig:moreegs} below. Fig \ref{fig:moreegs}(b) shows how diversifying on an axis like Geography can help diversify the Ethnicity distribution. 

\subsection{Validating Mitigation Effect Estimation}
\label{sup:mitigation}

Our approach provides empirical estimates of how a counterfactual-based mitigation strategy may influence an intersectional relationship $B_x \to B_y$ in the form of the \isscore\ score. To validate these estimates, we conduct an experiment where we actually perform mitigation on SD 1.4 using ITI-GEN. For all 26 occupations, we consider all intersectional relationships $B_x \to B_y$, and mitigate all $B_x$ independently. To compute the new \isscore\ post mitigation, we replace the initial distribution $D_{B_y}^{B_x}$ in equation 3 with $D_{B_y}^{mit(B_x)}$, which is based on the VQA extracted attributes for bias axis $B_y$ in the newly generated set of images post-mitigation of axis $B_x$ with ITI-GEN. This new score can be defined as:

\begin{equation}
w_{B_y}^{B_x} = W_1(D_{B_y}^{mit(B_x)}, D^*)
\end{equation}
\begin{equation}
IS_{xy}^{mit(x)} = w_{B_y}^{\text{init}} - w_{B_y}^{B_x} 
\end{equation}

Note that these equations are the same as the ones we used in Supp. \ref{sup:intro_study}, with the main difference being that, in this case, we only consider the scores for the intersectional relationships $B_x \to B_y$ found through causal discovery. To quantify the effectiveness of \modelname\, we measure the average correlation between the \isscore\ scores before $IS_{xy}$ and after mitigation $IS_{xy}^{mit(x)}$ across all intersectional relationships $B_x \to B_y$ present for each prompt.

In Table \ref{tab:mitigationcorr} in the main paper, we show these findings, and highlight that we achieved an average correlation of $+0.696$, suggesting that our method effectively estimates the potential impacts of bias interventions without actually doing the mitigation step itself, which often requires finetuning some or all parts of the diffusion model.

Such empirical guarantees provide users with valuable insights into whether altering bias along a particular dimension will lead to meaningful improvements in fairness across other bias dimensions. By estimating how counterfactual-based interventions influence overall bias scores, our approach helps researchers and practitioners predict the effectiveness of mitigation techniques before full deployment.

\subsection{Global Aggregations}
\label{sup:aggregation}

In order to do a comparative analysis of intersectionality across models over a dataset of prompts, we perform an aggregation step. For the 26 occupation prompts, we first start by using counterfactuals and VQA to identify attributes over all bias axes in $B$. Now, in the Causal Discovery step, we build contingency tables that aggregate attributes over all $CF$ prompts across all the occupations. For example, when considering the intersectional relationship $Gender \to Age$, we consider all images for \texttt{male} \textit{occupation} and \texttt{female} \textit{occupation} for all occupations for the rows of the contingency matrix, and count over the $Age$ attributes \texttt{young, middle-aged, old} to find the overall global distribution. This gives us the global contingency table for any bias pair. We follow the steps in Sec. \ref{sec:Approach:CausalDiscovery} to obtain this list of bias intersectionality relationships that are significant. Next, in order to compute \isscore, we use the same contingency table and sum across its columns to get $D_{B_y}^{global(B_x)}$. For the initial distribution, we accumulate attributes across all initial prompt images for all occupations, to give us $D_{B_y}^{global(init)}$. We can now compute \isscore\ as:

\begin{equation}
w_{B_y}^{global(\text{init})} = W_1(D_{B_y}^{global(\text{init})}, D^*)
\end{equation}
\begin{equation}
w_{B_y}^{global(B_x)} = W_1(D_{B_y}^{global(B_x)}, D^*)
\end{equation}
\begin{equation}
IS_{global(xy)} = w_{B_y}^{global(\text{init})} - w_{B_y}^{global(B_x)} 
\end{equation}

Given the large number of images (as we aggregate over multiple sets), we choose to use a p-value threshold of $0.00005$, and we further discard edges in the pairwise causal graph where the $-0.03 > IS_{global(xy)} > 0.03$.

\begin{figure}
  \centering
   \includegraphics[width=0.6\linewidth]{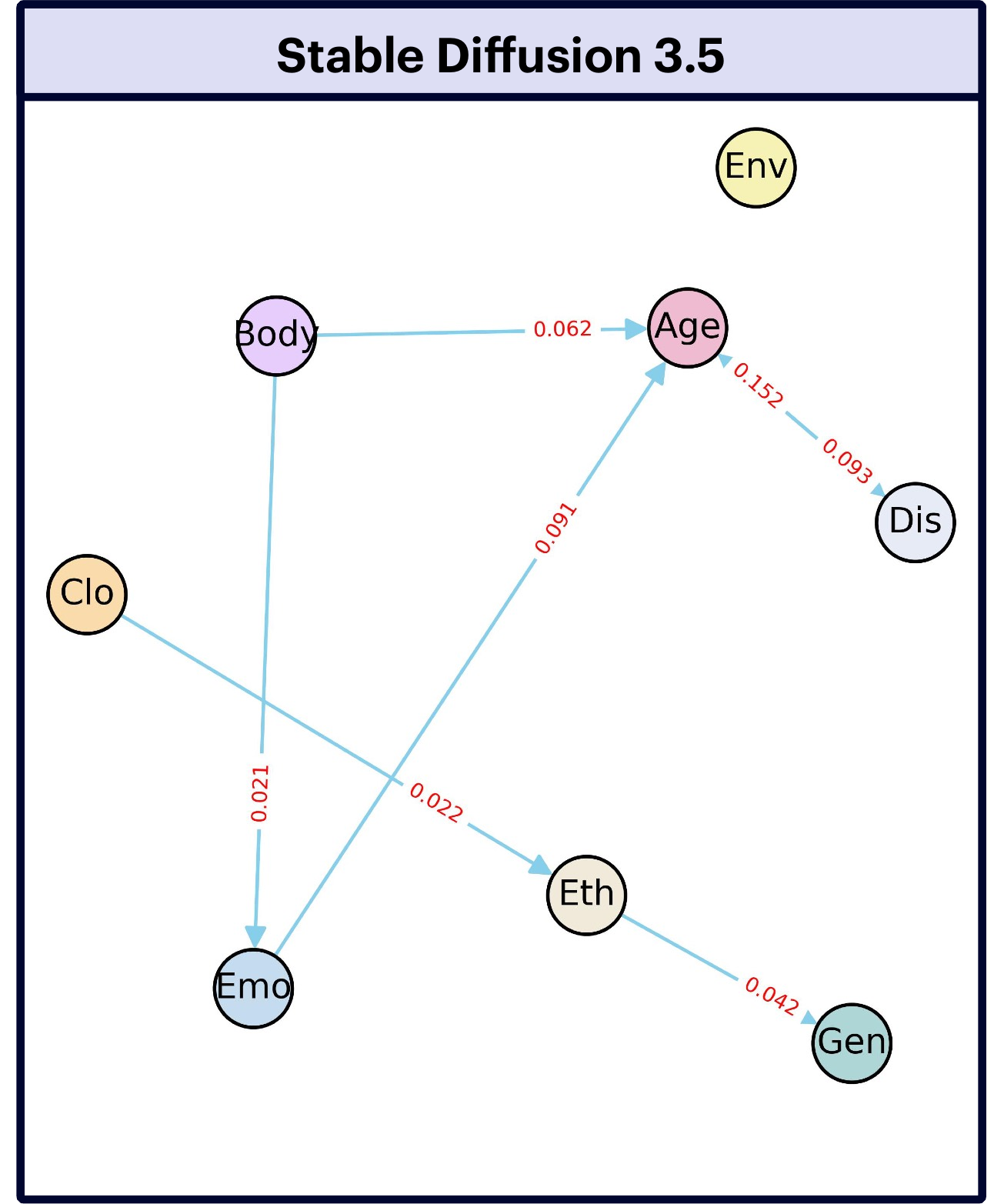}
   \caption{Global graph for Stable Diffusion 3.5. This graph is not included in the main paper due to space constraints. }
   \label{fig:sd35global}
\end{figure}

\subsection{Studying Real World Biases}
\label{sup:real_world}

\modelname\ can be used to compare bias dependencies in images generated by Text-to-Image (TTI) models with a reference real-world image distribution. Instead of assuming a uniform distribution as the baseline for bias sensitivity calculations, we consider the empirical distribution of the reference dataset as the initial distribution.

Given a prompt $P$ (e.g., ``A computer programmer''), let $B = [B_1, B_2, ..., B_n]$ represent the set of bias axes (e.g., gender, age, race). For each bias axis $B_y$, we define:
\begin{itemize}
    \item $D_{B_y}^{\text{real}}$: real-world distribution of $B_y$ (from a dataset or observed statistics).
    \item $D_{B_y}^{\text{TTI}}$: distribution of $B_y$ in TTI-generated images.
\end{itemize}

The Wasserstein-1 distance  between real-world and TTI-generated distributions quantifies how far the TTI bias distribution is from real-world data is:
\begin{equation}
    w_{B_y}^{\text{init}} = W_1(D_{B_y}^{\text{TTI}}, D_{B_y}^{\text{real}})
\end{equation}

To measure the impact of intervening on $B_x$, we compute the post-intervention Wasserstein distance:
\begin{equation}
    w_{B_y}^{B_x} = W_1(D_{B_y}^{B_x}, D_{B_y}^{\text{real}})
\end{equation}
The \isscore\ Score $IS_{xy}$ for the effect of changing $B_x$ on $B_y$ measures the difference between $w_{B_y}^{\text{init}}$ and $ w_{B_y}^{B_x}$ similar to the one calculated in Eq \ref{eqn:metric}.
To measure overall intersectional bias amplification, we compute:
\begin{equation}
    \mathcal{I} = \sum_{x \neq y} |IS_{xy}|
\end{equation}
where a high $\mathcal{I}$ indicates strong intersectional bias amplification, while a low $\mathcal{I}$ suggests minimal entanglement.

\begin{figure*}
  \centering
   \includegraphics[width=0.6\linewidth]{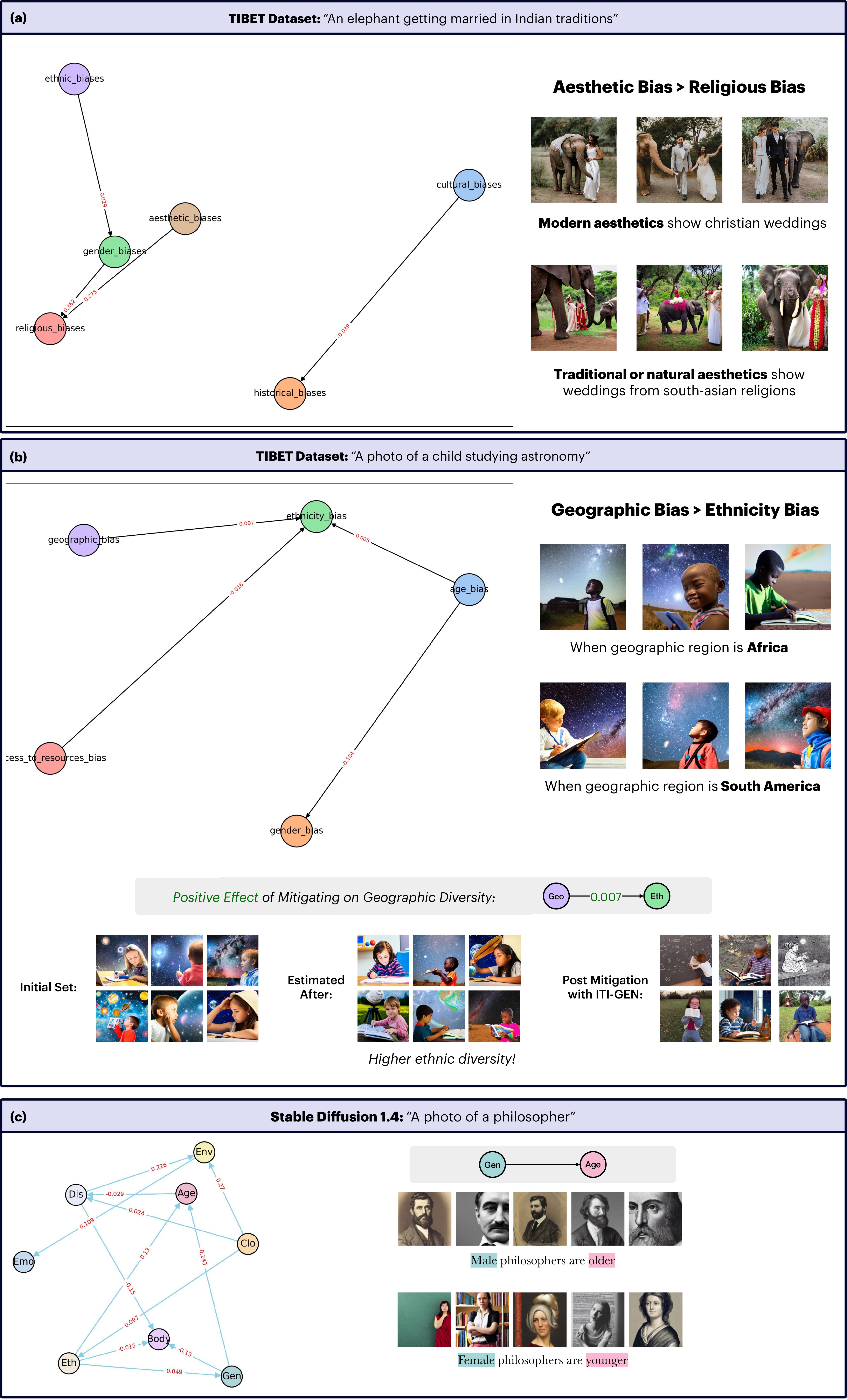}
   \caption{Additional examples on TIBET (a-b) and Occupation prompt (c) on prompt-level analysis provided by \modelname. }
   \label{fig:moreegs}
\end{figure*}

\end{document}